\documentclass[a4paper,fleqn]{cas-dc}

\usepackage[round,authoryear]{natbib}
\usepackage{subfigure}
\usepackage{amsmath}
\usepackage[ruled]{algorithm2e}
\usepackage{soul}
\usepackage{bm}
\usepackage{color, xcolor}
\usepackage{tabularx}
\usepackage{threeparttable}
\usepackage{chngcntr}
\usepackage{ltablex}
\usepackage{multirow}
\usepackage{caption}
\usepackage{fancyhdr}
\usepackage{lastpage} 
\soulregister\cite7
\soulregister \citep7
\soulregister\ref7
 
\counterwithout{table}{section} 
\keepXColumns 

\def\tsc#1{\csdef{#1}{\textsc{\lowercase{#1}}\xspace}}
\tsc{WGM}
\tsc{QE}

\fancypagestyle{plain}{%
  \fancyhf{} 
  \rfoot{Page \thepage\ of \pageref{LastPage}} 
}

\begin{document}
\let\WriteBookmarks\relax
\def\floatpagepagefraction{1}
\def\textpagefraction{.001}


\title [mode = title]{ESGReveal: An LLM-based approach for extracting structured data from ESG reports}



%

\author{Yi Zou $^{a,1}$}
\address[1]{Alibaba Cloud, Hangzhou, China}
\author{Mengying Shi $^{b,1}$}
\address[2]{Department of Earth System Science, Tsinghua University, Beijing, China}



\author{Zhongjie Chen $^{a}$}   
\author{Zhu Deng $^{a}$}
\author{ZongXiong Lei $^{a}$}
\author{Zihan Zeng $^{c}$}
\address[3]{Department of Environmental Science and engineering, Sun Yat-Sen University, Guangzhou, China 
\\$^1$These authors contribute equally.}
\author{Shiming Yang $^{c}$}
\author{HongXiang Tong $^{a}$}
\author{Lei Xiao $^{a}$}
\author{Wenwen Zhou $^{a}$}
\ead{zhoubo.zww@alibaba-inc.com}
\cormark[1]

\cortext[1]{Corresponding author.}



\begin{abstract}
ESGReveal is a method introduced in this paper for the systematic extraction and analysis of Environmental, Social, and Governance data from corporate reports, designed to address the pressing need for consistent and accurate retrieval of ESG information. Using  Large Language Models (LLM) combined with Retrieval Augmented Generation (RAG) techniques, ESGReveal includes an ESG metadata module for criteria queries, a report preprocessing module for building databases, and an LLM agent module for data extraction. The framework's effectiveness was evaluated using ESG reports issued by companies across 12 industries listed on the Hong Kong Stock Exchange  in 2022. A carefully selected representative sample of 166 companies, based on industry distribution and market capitalization, provided a comprehensive assessment of ESGReveal's capabilities. The application of ESGReveal yielded significant findings on the current state of ESG reporting, with GPT-4 achieving accuracy rates of 76.9\% in data extraction and 83.7\% in disclosure analysis, outperforming baseline models. These results suggest the framework's utility in improving the accuracy of ESG data analysis. Additionally, our analysis identified the need for more robust ESG practices, with environmental data disclosure at 69.5\% and social data at 57.2\%, indicating room for increased corporate transparency. Recognizing the current limitations of ESGReveal, including its inability to interpret pictorial data which is a feature planned for future enhancement, the study also suggests additional research to improve and compare the differential analytical performance of various LLMs. In conclusion, ESGReveal marks a step forward in ESG data processing, providing stakeholders a tool to better assess and enhance corporate sustainability practices. Its development shows promise in advancing the transparency of corporate reporting and contributing to the wider objectives of sustainable development.

\end{abstract}


\begin{keywords}
  ESG \sep Data Extraction  \sep Large Language Models\sep ChatGPT
\end{keywords}
\maketitle
\thispagestyle{plain}





\section{Introduction}
Since the United Nations Global Compact initiated the concept of Environmental, Social, and Governance (ESG) in 2004, corporations have utilized ESG reporting to show their initiatives and commitment within these domains \citep{tsang2023esg}. Across the globe, myriad ESG reporting frameworks, notably the Global Reporting Initiative (GRI) and the Sustainability Accounting Standards Board (SASB), have gained widespread acceptance. Additionally, numerous stock exchanges have enacted ESG disclosure directives to guide corporate reporting practices.

ESG disclosure quantifies corporate transparency and is crucial for assessing performance in ESG aspects, providing a basis for decision-making for investors and other stakeholders \citep{bui2020climate}. While investors and analysts can access ESG reports through stock exchanges and corporate websites, the vast number and varied formats of these reports pose significant integration challenges for consolidating disclosure data at the corporate or industry level \citep{doe2021glitter}. Third-party ratings such as those from MSCI, Sustainalytics, and Bloomberg help understand corporate non-financial performance but lack the necessary transparency and detailed disclosure metrics \citep{abhayawansa2021sustainable,schiemann2022esg}. Research indicates that there is currently no publicly available ESG disclosure database covering detailed metrics, which limits the depth of analysis and regulation of corporate ESG performance to some extent \citep{clarkson2019causes}.

In response to these impediments, this study introduced an approach, designated as ESGReveal, based on the integration of advanced Large Language Models (LLM) and Retrieval Augmented Generation (RAG) techniques. The ESGReveal is comprised of a tripartite architecture: ESG metadata module, report preprocessing module, and LLM agent module. Furthermore, an analytical exploration utilizing ESG reports from preeminent corporations across sectors listed on the Hong Kong Stock Exchange in year 2022 was conducted. The main contributions of this research include:

$\bullet$  Designing and executing ESGReveal for systematic extraction of crucial numerical and textual data from corporate ESG reports.

$\bullet$  Assessing the performance of different large language models in ESG information retrieval, setting a baseline for further ESG data processing and analytical studies.

$\bullet$  Utilizing ESGReveal to evaluate ESG reports from a representative subset of companies on the HKEx, providing industry benchmarks for ESG conduct and reporting.

\begin{figure*}[!t]
\centering
\includegraphics[width=.95\linewidth]{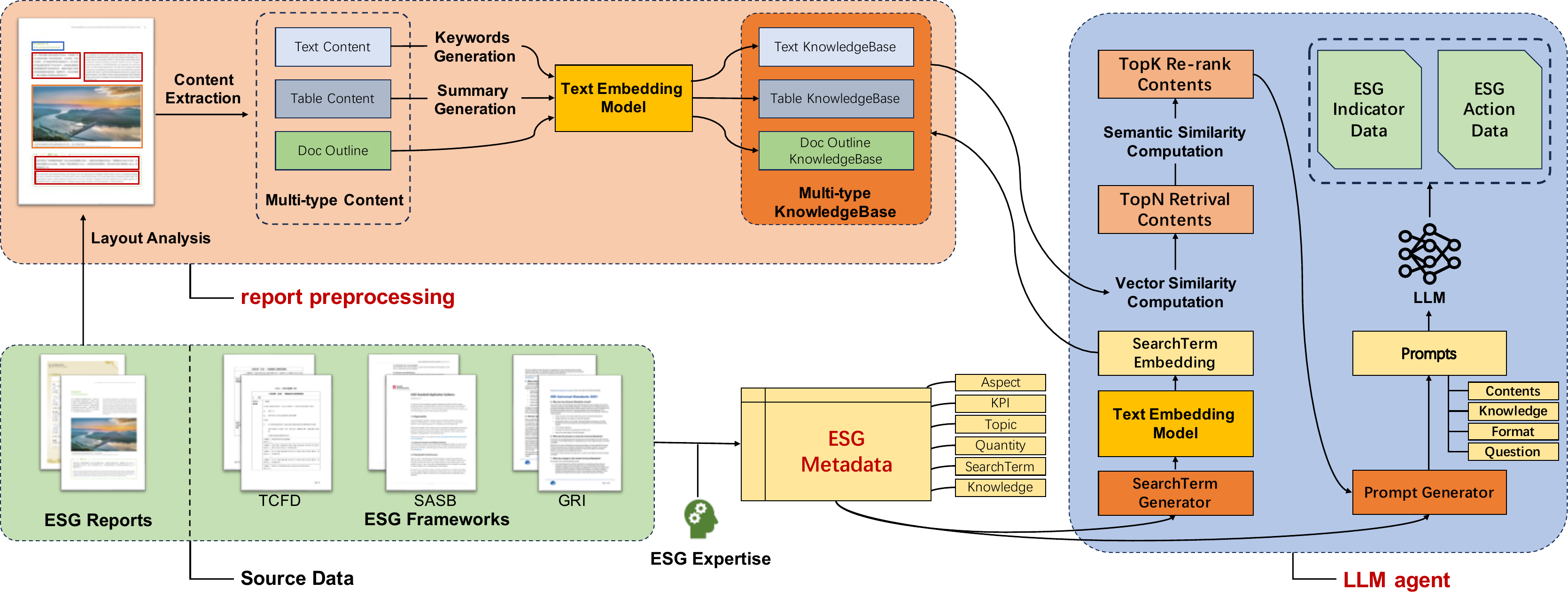}
\caption{\textrm{Overall structure of ESGReveal.}}
\label{fig:1}
\end{figure*}

\section{Related Work}
In the face of the complexity of ESG disclosure and the rapid growth in information volume, researchers have begun to explore the potential of natural language processing (NLP) techniques to enhance the precision and efficiency of ESG data extraction and analysis. \cite{ruberg2023greenai} utilized the BERT architecture to automatically classify content in corporate ESG reports relevant to the GRI standards, increasing the analytical efficiency for ESG assessments at the Brazilian Development Bank. \cite{perazzoli2022evaluating} conducted an extensive analysis of 55,000 publications on ESG-related topics using NLP techniques. \cite{luccioni2020analyzing} developed the ClimateQA model based on NLP technology to analyze information related to climate change in financial reports.\cite{raman2020mapping} detected ESG trends by analyzing corporate earnings call transcripts and parsing the linguistic structure within ESG texts. \cite{fischbach2022automatic} created a tool named ESG-Miner, which automatically extracts ESG-related information from media headlines and calculates ESG scores for companies. In these studies, advanced NLP models like BERT have made significant progress in understanding ESG contexts and semantics \citep{liu2023public,lee2022proposing}, yet there remain notable limitations in areas such as adaptability to the ESG sector, custom datasets, deep information mining, and multilingual and cross-cultural adaptability \citep{pasch2022nlp,fischbach2022automatic}.

In the field of natural language processing, recent developments in large language models such as GPT-3.5 \citep{openai2023instructgpt} and GPT-4 \citep{openai2023gpt4}. To address challenges with lengthy contexts and data currency in LLM, researchers have integrated RAG paradigm \citep{lewis2020retrieval} with LLM, which improves response precision with dynamic information retrieval. Current research reveals the substantial potential of LLM in identifying sustainability goals and performance indicators within ESG reports \citep{zhang2023renovation,burnaev2023practical}. For instance, \cite{kim2023bloated} used ChatGPT to summarize the economic utility of corporate disclosures and revealed the link between information "bloat" and adverse capital market consequences. \cite{ni2023chatreport} developed the CHATREPORT system based on LLM, which analyzed 1015 corporate sustainability reports using the Task Force on Climate-Related Financial Disclosures (TCFD) framework (https://www.fsb-tcfd.org/) and assessed their compliance. \cite{bronzini2023glitter} extracted semantically structured ESG-related information from sustainability reports using LLM, unveiling inter-corporate ESG action correlations. \cite{moodaley2023conceptual} enhanced the ability to identify green claims and detect "greenwashing" by training LLM with sustainability-related textual corpora. \cite{yang2023fingpt}’s FinGPT showed promising application potential in recognizing information pertinent to environmental, social, and governance issues. The evolution of LLM has greatly optimized text processing methods, making the transformation of unstructured documents in ESG reports into structured data a feasible reality \citep{visalli2023esg}.

Building on the foundation of existing research, this paper presents ESGReveal, a novel methodology that focuses on the extraction of data from ESG reports with a particular emphasis on the structured retrieval of numerical information. Our approach harnesses the sophisticated text processing capabilities of LLMs within RAG paradigm to systematically decompose both numerical and textual information contained in ESG reports from various standard frameworks. This systematic analysis facilitates deeper insights into ESG disclosure practices.

\begin{figure*}[!t]
\centering
\includegraphics[width=.95\linewidth]{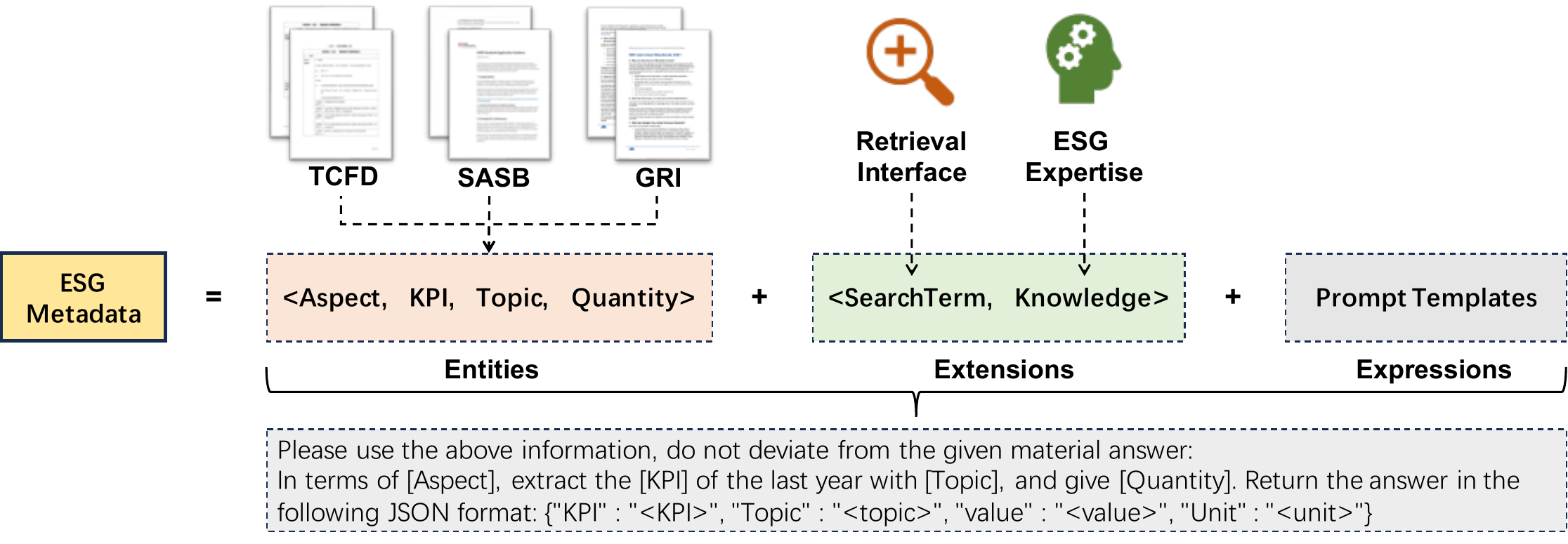}
\caption{\textrm{Structure of ESG metadata module: Entities, Extensions, and Expressions.}}
\label{fig:2}
\end{figure*}

Furthermore, in recognition of the pivotal role played by the Hong Kong Stock Exchange (HKEx) in advancing ESG initiatives in China, this study employs ESGReveal to evaluate the ESG reports of companies listed on the HKEx. Through this application, we aim to analyze the prevailing state of industry transparency and accountability in ESG, contributing towards a clearer understanding of the ESG landscape within the region.

\section{Methods}
In this section, we describe the architecture and components of ESGReveal. The general framework of ESGReveal is introduced in subsection 2.1. Following that, subsections 2.2, 2.3, and 2.4 elaborate on the primary modules: ESG metadata module, report preprocessing module, and LLM agent module.

\subsection{Overall}
Figure \ref{fig:1} illustrates the three module design of ESGReveal. The ESG metadata module establishes a query framework that leverages ESG criteria and expertise to analyze reports. The report preprocessing module imports ESG reports and processes them to build a database for information retrieval. The LLM agent module then accesses this database to retrieve information and engages the LLM for data extraction, guided by the ESG metadata module In the ES9Reveal workflow, an ESG report is processed by the report preprocessing module and analyzed by the LLM agent module, resulting in data that adheres to the ESG specifications defined by the ESG metadata module.

\subsection{ESG Metadata Module}
 ESG metadata module is a metadata framework structured to comply with international ESG standards, such as GRI and SASB, and is segmented into three components: Entities, Extensions, and Expressions, as depicted in Figure \ref{fig:2}. Entities outline the attributes of an indicator, detailing its category, description, subcategories, and measurement units. Extensions encompass <Knowledge> and <SearchTerm> to adapt to various ESG compilation standards. Expressions concentrate on crafting Prompt templates and formatting the output, which allow for the automatic generation of prompts to trigger the LLM. Subsequently, this study detailed ESG metadata module’s Entities and Extensions based on the \textit{Supplementary 27 Environmental, Social and Governance Reporting Guide} \citep{hongkongexchanges2023esg} and the \textit{Supplementary 14 CORPORATE GOVERNANCE CODE} \citep{hongkongexchanges2023corporate} published by HKEx.

\subsubsection{Entities}
The ESG metadata module’s Entities simplify complex ESG issues into the format: <Aspect, KPI, Topic, Quantity>, with "KPI" denoting key performance indicators. Adhering to HKEx guidelines, the framework comprises 70 indicators, divided as 12/18/4 numerical and 14/15/7 textual indicators across the E/S/G categories, totaling 34 numerical and 36 textual indicators—detailed in Table \ref{tab:1} (for details, see Supplementary Table \ref{supplefig:1}) . Examples of Entities are as follows (for details, see Supplementary Table 2) :

Example 1: On the "A1. Emissions" issue, for querying total and intensity of waste emissions, <Aspect, KPI, Topic, Quantity> are "A1. Emissions", "Total waste produced (in tonnes) and, where appropriate, intensity", "Non-hazardous Waste, Hazardous Waste…" and "Absolute Values"

Example 2: On "A1. Emissions" issue, regarding key actions on emissions, <Aspect, KPI, Topic, Quantity> are "A1. Emissions", "Emissions target(s) and steps taken to achieve them", "Waste, Exhaust/Greenhouse Gases…", and "Key Actions"


\setcounter{table}{0}
\begin{table*}[ht]
\centering
\caption{\textrm{Indicators of the ESG Metadata Module Based on HKEx Standards. The Full List of ESG Indicators Can Be Found in Supplementary Table 1.}}
\includegraphics[width=0.95\textwidth]{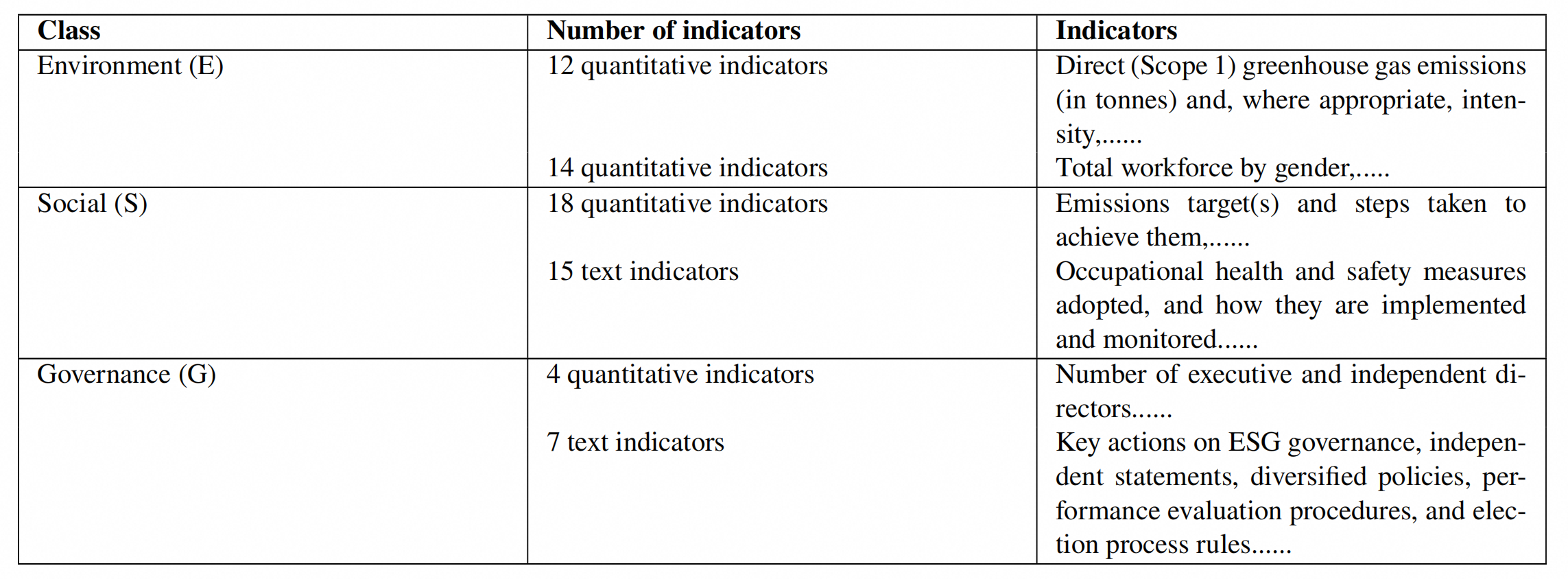}
\end{table*}

\subsubsection{Extensions}
To enhance adaptability across diverse ESG reporting standards, <Knowledge> and <SearchTerm> are integrated into ESG metadata module. The <Knowledge> component, containing domain knowledge curated by ESG experts and user-friendly explanations of indicators produced by sophisticated LLM, is embedded into LLM prompts via In-Context Learning to accurately carry out indicator extraction tasks. The <SearchTerm> component includes customized retrieval keywords gathered, for instance, from an extensive array of ESG reports based on the HKEx guidelines to improve the performance of the RAG retrieval. Illustrations of <Knowledge> and <SearchTerm> are delineated in Table \ref{tab:2}, with comprehensive elaboration presented in Supplementary Table \ref{suppletab:2}.

\subsection{Report Preprocessing Module}
To enhance the effectiveness of ESG indicator extraction, we have specially optimized the preprocessing procedures for ESG reports and the construction of the knowledge base.

\subsubsection{Preprocess of ESG Report}
In the preprocessing of ESG reports, we initially employed advanced computer vision tools, including Microsoft's LayoutLMv3 \citep{huang2022layoutlmv3} and GeoLayoutLM \citep{xing2023lore}, to extract the structural components of documents, such as headers, paragraphs, and tables. Subsequently, we conducted structural analysis using font characteristics from structural components to construct a report outline that connects to the text, thereby facilitating quicker information retrieval. Finally, recognizing the prevalence of numerical indicators in tables, we implemented algorithms like Microsoft's Table-Transformer \citep{smock2022pubtables} and LORE-TSR \citep{xing2023lore2} to accurately identify table cells and reconstruct table structures, ensuring the precise representation of tabular data.

\subsubsection{Construction of Multi-type Knowledge Base}
After preprocessing, we organized ESG reports data into structured knowledge bases for textual contents, document outlines and table contents. For textual contents, we generated summaries using the model mt5 \citep{raffel2020exploring} and created vector representations with the model m3e \citep{wang2023m3e}, storing them in vector databases like FAISS \citep{johnson2019billion} and Milvus \citep{wang2021milvus}. Document outlines were treated similarly, linking information with corresponding vectors for storage. For table contents, we employed a one-to-many mapping to exclude non-essential details and highlight key information such as ESG indicator names. We then generated vector representations for these keywords, pairing each table with its vector list to establish a concise knowledge base for table data.


\setcounter{table}{1}
\begin{table*}[ht]
\centering
\caption{\textrm{Contents of the ESG Metadata Module. The Full List of ESG Metada Can Be Found in Supplementary Table 1.}}
\includegraphics[width=0.95\textwidth]{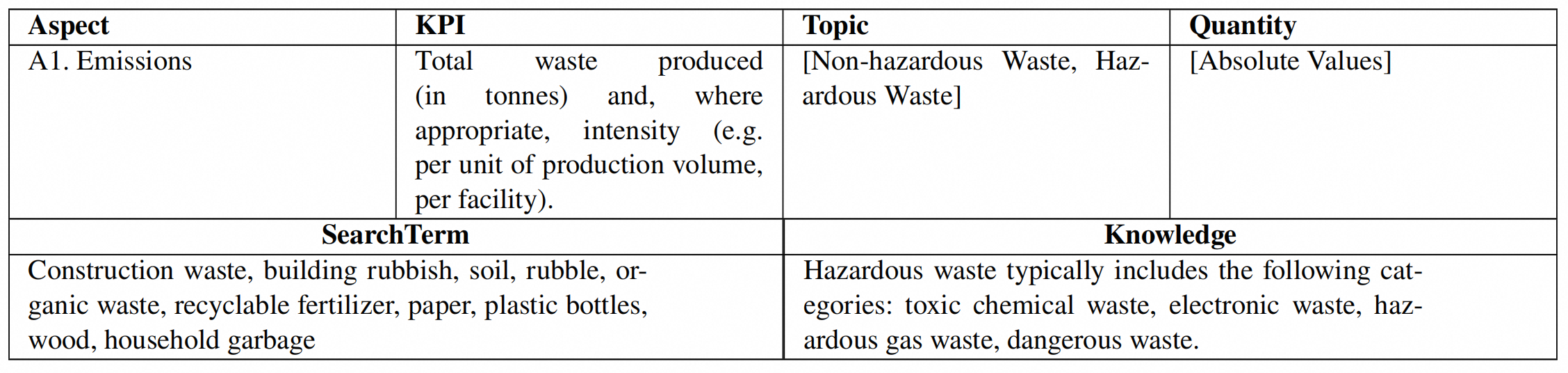}
\end{table*}

\subsection{LLM Agent Module}
\subsubsection{Retrieval of Knowledge}
Following processing by the report preprocessing module, we enabled precise ESG metadata module-driven searches for ESG indicator-specific data from knowledge bases. We created vectors for queries generated by metadata using the same methods employed by the report preprocessing module and calculated their cosine similarity to those in the knowledge base. Vector retrieval helped us pinpoint the most pertinent findings from text, tables, and outlines. To improve retrieval accuracy and relevance, we applied the coROM \citep{zhang2022hlatr}  model to assess the semantic similarity of the initial top matches, which allowed us to refine their order. Ultimately, we extracted the entries with the greatest similarity for reliable ESG indicator-specific knowledge.

\subsubsection{LLM Answering}
After data retrieval, we created a prompt based on ESG metadata module and retrieved contents to extract indicator information. The prompt comprises the following elements: (1) Preset Information, which provides basic instructions for the LLM's behavior; (2) Reference Content, which includes the retrieved contents from the knowledge base; (3) Expert Knowledge, which incorporates ESG insights from specialists into ESG metadata module; (4) Question, which frames a targeted query for the indicator crafted from ESG metadata module, asking about the indicator's disclosure and related data; (5) Answer Format, which presents a structured request for the LLM to report data in the format of <Disclosure, KPI, Topic, Value, Unit, Target, Action>, covering disclosure status, ESG performance indicators, topics, numerical values, units, action targets, and key actions, respectively. For detailed prompt construction and response examples, see Supplementary Table \ref{suppletab:3}

\section{Datasets}
We collected approximately 2249 ESG reports from 2022 issued by companies listed on HKEx (https://www.hkex.
com.hk/), categorized into 12 industries according to the Hang Seng Industry Classification System \citep{hongkongexchanges2023esg}. As processing all ESG reports entailed a significant computational workload, we endeavored to select a representative sample of 166 companies, balancing industry diversity and market capitalization, for evaluating ESGReveal and performing disclosure analysis. ESG datasets for different industries are shown in Table \ref{tab:3}. For more details about industries and companies, please refer to Supplementary Table \ref{suppletab:4}.

\section{Results}
\subsection{Assessing ESGReveal: Accuracy Indicators, LLM Benchmarking and Ablation Study}
ESG disclosure frameworks encompass a wide range of reporting indicators, broadly categorized into quantitative (e.g., greenhouse gas emissions) and qualitative (e.g., employee diversity policies). Due to difficulties in quantifying the latter, this study focused on the precision of ESGReveal in extracting quantitative indicators as specified by ESG metadata module. We detailed the accuracy assessment method in Section 5.1.1, compared various LLMs' performances in Sections 5.1.2, and discussed ablation study results using the ESG metadata module framework in Section 5.1.3. This study utilized the ESGReveal tool to dissect selected sample reports, extracting critical information on ESG data disclosures, quantitative figures, and key initiatives, with detailed findings presented across sections 5.2.1 to 5.2.3.


\setcounter{table}{2}
\begin{table*}[ht]
\centering
\caption{\textrm{Sectors and firms of ESG reports. The Full List of ESG Reports Can Be Found in Supplementary Table 4.}}
\includegraphics[width=0.95\textwidth]{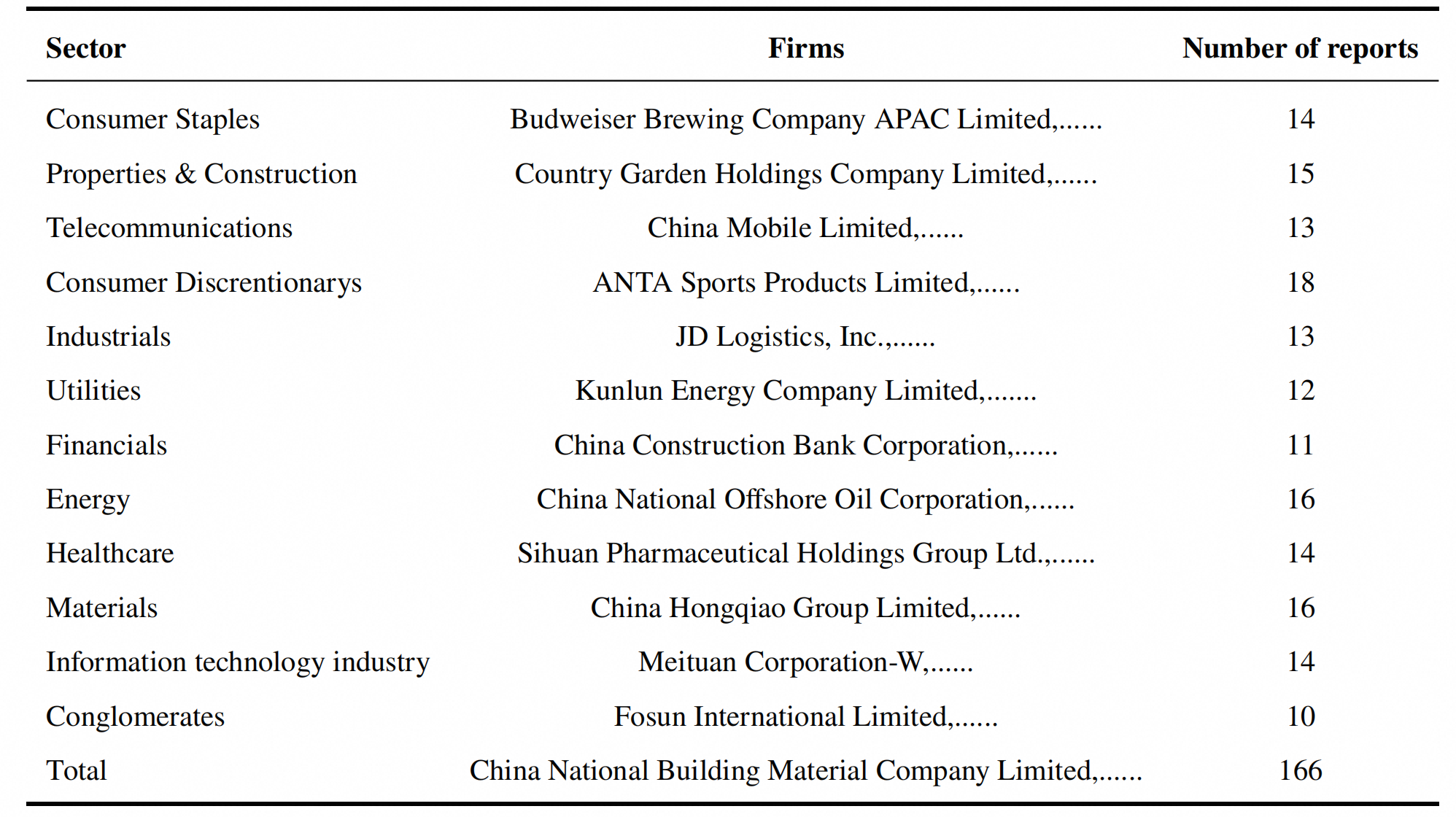}
\end{table*}

\subsubsection{Accuracy Indicators}
Utilizing ESGReveal, this study extracted the degree of indicator disclosure and their specific numerical values from sample reports under the HKEx framework. To quantitatively assess the accuracy of data extraction, we calculated the disclosure coverage accuracy ($Acc_{DC}$) and the data extraction accuracy ($Acc_{DE}$).
Initially, we manually annotated the numeric indicators within the sample reports to obtain the disclosure status $D_{label}$ and specific values $V_{label}$ for each company's indicators.

\begin{equation}\label{eq:1}
D_{\text{label}} = \{d_i | i \in \{1,2,3,\ldots,N_{\text{mq}}\}, d_i \in \{1,0\}\}
\end{equation}

\begin{equation}\label{eq:1}
V_{\text{label}} = \{v_j | j \in \{1,2,3,\ldots,N_{\text{v}}\}\}
\end{equation}

Here, $N_{mq}$ represents the total number of indicators defined in ESG metadata module, with the disclosure status of each indicator $d_i$ being marked as "1" (disclosed) or "0" (not disclosed). $N_{v}$ denotes the total number of numeric indicators that have been actually disclosed, with the specific value of each indicator $V_{j}$ represented in JSON format.

Subsequently, $Acc_{DC}$ is defined as the ratio of the number of disclosure indicators correctly identified by ESGReveal to the number of indicators actually disclosed in the reports. Similarly, data extraction accuracy $Acc_{DE}$ is defined as the ratio of the number of numeric values correctly output by ESGReveal to the number of true numeric values in the sample reports.

\begin{equation}\label{eq:1}
\begin{aligned}
    Acc_{DC}=\
    \frac{1}{N_{mq}} \sum_{i=1}^{N_{mq}} 1(d_i=\hat{d_i}) \text{ ,}\\
    \text{ where }  d_i\in D_{label}  \text{ and }  \hat{d_i} \in D_{ESGReveal}
\end{aligned}
\end{equation}

\begin{equation}\label{eq:1}
\begin{aligned}
    Acc_{DE}=\
    \frac{1}{N_{v}} \sum_{j=1}^{N_{v}} 1(v_j=\hat{v_j}) \text{ ,}\\
    \text{ where }  v_j\in V_{label}  \text{ and }  \hat{v_j} \in V_{ESGReveal}
\end{aligned}
\end{equation}

Here, $D_{ESGReveal}$ and $V_{ESGReveal}$ represent the disclosure status of indicators and specific values extracted by ESGReveal, respectively.

These accuracy metrics are calculated as the average across the studied samples in order to evaluate the overall performance of ESGReveal.

\subsubsection{LLM Benchmarking}
Table \ref{tab:4} shows the performance of ESGReveal when extracting ESG data using different LLMs. For comparative analysis, four advanced models were chosen: GPT-3.5, GPT-4, ChatGLM \citep{du2022glm}, and the QWEN \citep{bai2023qwen} model.

The experimental results indicate that compared to $Acc_{DC}$, all LLMs generally showed a decline of 4.8\% to 7.7\% in $Acc_{DE}$. This consistent performance gap may reflect the inherent complexity of the data extraction task, which requires not only identifying the presence of disclosed information but also accurately extracting specific values and attributes. Particularly when dealing with the extraction of data on topics such as "Employee turnover rate by gender, age group and geographical region" the task involves detailed values for multiple age brackets. This requires that the LLM be capable of accurately processing and distinguishing between multiple data points.

In the performance comparison of different LLMs, GPT-4 achieved an accuracy of 76.9\% in $Acc_{DE}$ and 83.7\% in $Acc_{DC}$, leading other models. QWEN followed with accuracy of 54.9\% and 61.4\%, respectively. Lastly, GPT-3.5 and ChatGLM performed similarly, with GPT-3.5 achieving 47.1\%/51.9\% and ChatGLM achieving 46.2\%/53.9\% accuracy rates. The superiority of GPT-4 in the analysis of ESG disclosures and the extraction of relevant indicators can be attributed to its enhanced model capacity and superior comprehension capabilities. GPT-3.5, QWEN and ChatGLM have equivalent capabilities and are inferior to GPT-4. These findings underscore the pivotal role that a model's capacity and interpretive proficiency play in augmenting its effectiveness when applied to specialized domain-specific tasks.

\subsubsection{Ablation Study}
Based on the experimental performance in Section 5.1.2, we selected GPT-3.5 and GPT-4 as representatives to conduct a series of ablation experiments to evaluate the effectiveness of each module in ESGReveal proposed in this study. We used a basic RAG implementation as the Benchmark \citep{langchain2023}. As illustrated in Figure 3, under the Benchmark implementation, GPT-4 achieved accuracy of 57.8\% in disclosure analysis and 52.2\% in data extraction, while GPT-3.5 achieved 31.7\% and 26.6\%, respectively.

\textit{1) Effectiveness of the Enhanced RAG}

Building upon the Benchmark, we utilized the improved the report preprocessing module and LLM agent module, which significantly enhanced the performance of ESGReveal through better document preprocessing and content retrieval methods. As shown in Figure 3, marked as Enhanced-RAG, GPT-4 saw notable improvements in performance for both types of tasks, achieving 81.2\% (+23.4\%) for disclosure analysis and 74.0\% (+21.8\%) for data extraction. GPT-3.5 also brings improvement of 10.3\% and 12.3\%, reaching a performance of 42.0\% and 38.9\%, respectively. These results underscore the effectiveness of report preprocessing and LLM agent modules proposed in this study. Better structuring of documents and refined retrieval content significantly improve the performance of ESG disclosure analysis and data extraction under the RAG framework.

\textit{2) Effectiveness Analysis of ESG Knowledge}

With the introduction of <Knowledge> of ESG metadata module, both GPT-4 and GPT-3.5 achieved varying degrees of improvement. GPT-4’s accuracy increased to 83.7\% (+2.5\%) for disclosure analysis and 76.9\% (+2.9\%) for data extraction, whereas GPT-3.5 saw substantial increases of 9.9\% and 8.2\%, with accuracy reaching 51.9\% and 47.1\%, respectively. This indicates that by incorporating supplementary knowledge from the ESG domain, LLMs can experience different levels of enhancement in their parsing abilities. Especially for the LLMs with weaker model capacity and comprehension abilities (e.g., GPT-3.5), the accuracy gains are much greater than those for LLMs with stronger capacities and comprehension (e.g., GPT-4). This implies that the addition of ESG domain knowledge in ESG Metadata can greatly enhance the performance of less capable LLMs in ESG disclosure analysis and data extraction tasks.

\begin{figure}
    \centering
    \includegraphics[width=1\linewidth]{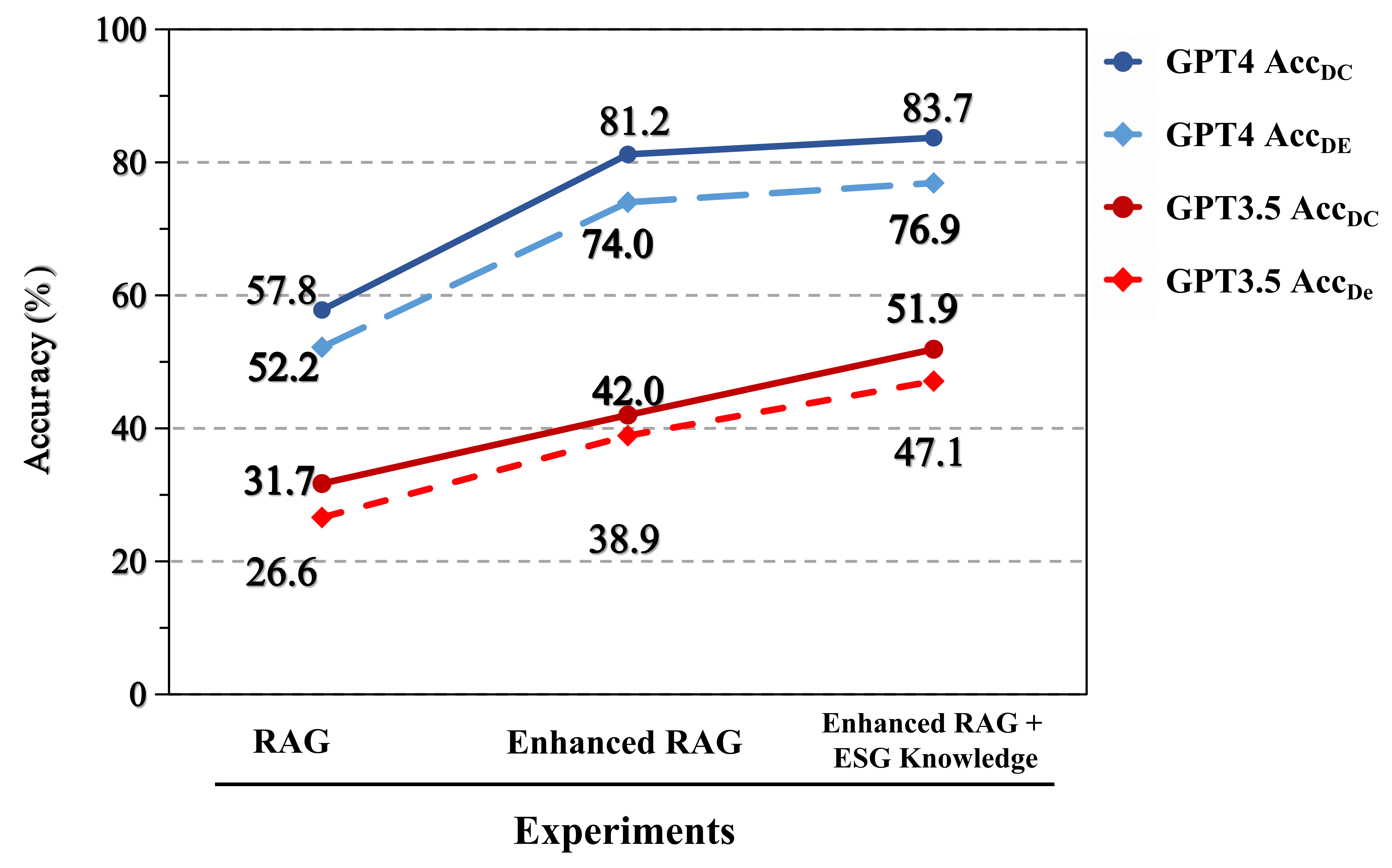}
    \caption{\textrm{Ablation Study of ESGReveal}}
    \label{fig:3}
\end{figure}

\subsection{Dissecting ESG Reports: Disclosure, Metrics, and Key Actions}
This study applied ESGReveal to parse selected sample reports, thereby obtaining information on ESG data disclosure, related quantitative data, and key actions within the reports. Subsequently, in Sections 5.2.1 to 5.2.3, we described and conducted a preliminary analysis of the extracted data, in order to understand the specific performance of the sample reports in terms of ESG aspects.

\subsubsection{Disclosure}
\begin{figure*}[!t]
\centering
\includegraphics[width=.95\linewidth]{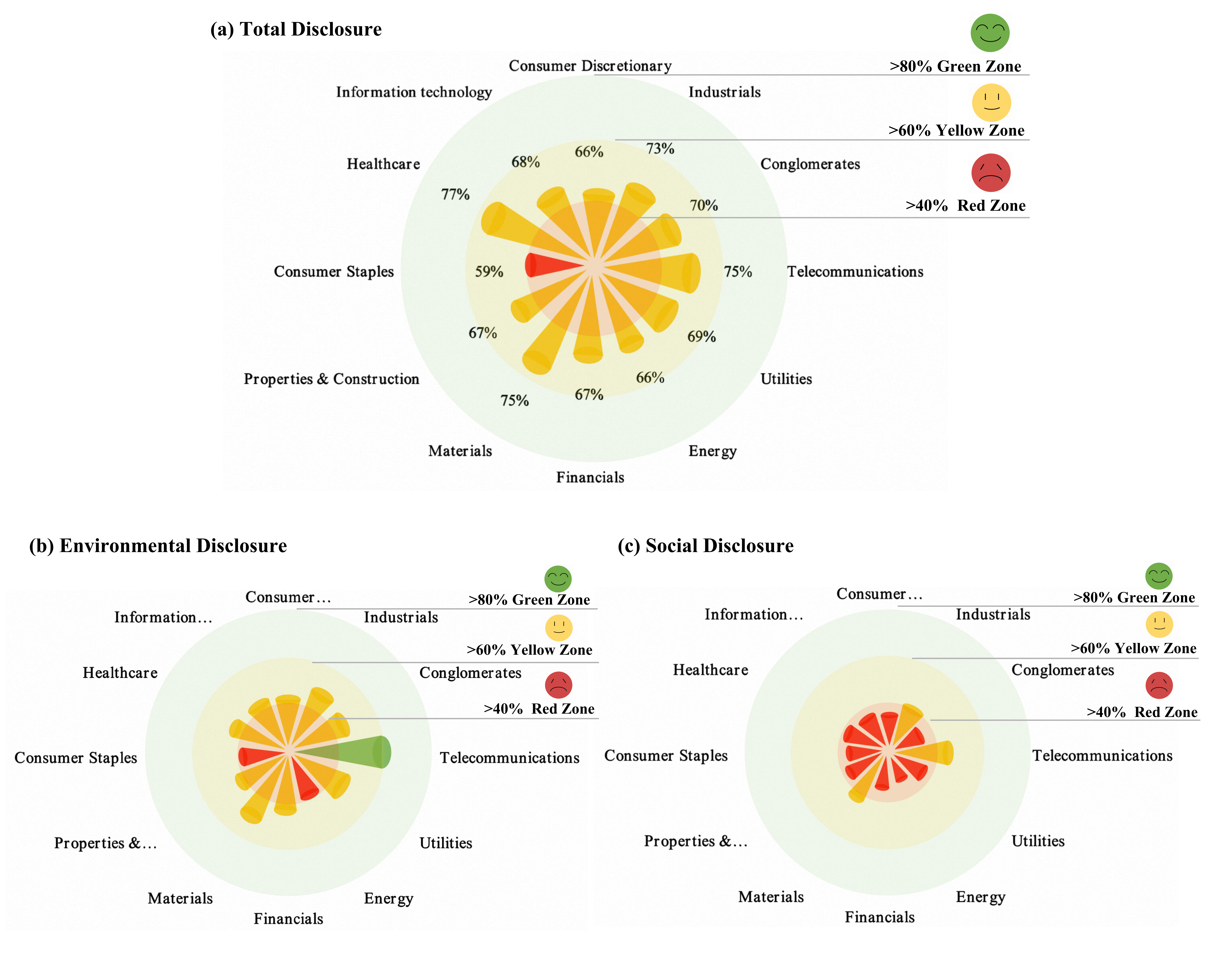}
\caption{\textrm{Comparative Analysis of Disclosure Levels by Industry: (a) Overall Disclosure, (b) Environmental Disclosure, and (c) Social Disclosure.}}
\label{fig:4}
\end{figure*}

Guided by the HKEx framework, this study designed 34 numerical indicators, specifically, the environmental dimension includes 12 indicators, while the social dimension involves 18 indicators. In the detailed analysis of corporate ESG reports, we also observed that the vast majority of companies tend to disclose data primarily on these two dimensions. For visual clarity, we display the disclosure of environmental, social, and overall numerical indicators across various industries in Figure \ref{fig:4}. Following the practice of professional ESG rating agencies like S\&P Global and MSCI, which categorize ESG rating results into high, medium, and low tiers based on score ranges \citep{spdjindices2023esg}, and in conjunction with the actual disclosure practices of various industries, this study adopted a three-tier assessment system to evaluate industry disclosure levels, namely: excellent (over 80\%), moderate (over 60\%), and poor (over 40\%).

\setcounter{table}{3}
\begin{table}[!ht]
\caption{${Acc_{DE}}$ and ${Acc_{DC}}$ Across LLMs.}
\label{tab:4} 
\centering
\renewcommand{\arraystretch}{1.5}
\begin{tabular}{lcccc}
    \toprule[1.2pt]
    \textbf{Models} & \textbf{GPT-4} & \textbf{QWEN} & \textbf{GPT-3.5} & \textbf{ChatGLM}\\
    \midrule
    $Acc_{DE}$       & 76.9\% & 54.9\% & 47.1\% & 46.2\% \\
    $Acc_{DC}$       & 83.7\% & 61.4\% & 51.9\% & 53.9\% \\
    \bottomrule[1.2pt]
\end{tabular}
\end{table}

The statistical results show that the average environmental disclosure rate is 69.5\%, while the average social disclosure rate is 57.2\%, indicating that, in general, environmental disclosure is superior to social disclosure. Moreover, even among the representative subset of companies by market capitalization listed on the HKEx, the average industry disclosure level has not exceeded 80\%. This finding suggests that there is an urgent need to strengthen ESG disclosure practices within the industry. Upon further examination of the healthcare industry, which possesses the highest disclosure level among all sectors, for industry analysis (see Supplementary Figure \ref{supplefig:1}). Within the healthcare sector, there are 5 companies that show good disclosure performance, while the remaining 6 are all rated as moderate. From the perspective of industry distribution, the telecommunications sector, such as China Tower Corporation Ltd., China Mobile Ltd., and China Unicom (Hong Kong) Ltd., healthcare sector, like China Biologic Products Holdings Inc., Alibaba Health Information Technology Ltd., and Materials sector, such as Northern Mining Ltd. stand out with higher disclosure rates, while the Consumer Staples, such as China Starch Holdings Ltd., Want Want China Holdings Ltd. and the energy sector, like China National Offshore Oil Corporation Ltd., China Petroleum \& Chemical Corporation Ltd. show weaker performance.

\subsubsection{Metrics}
In the process of parsing ESG reports with ESGReveal, we not only examined the disclosure status but more importantly, extracted quantitative data from the reports. Taking the data on Scope 1 emissions (direct emissions) and Scope 2 emissions (indirect emissions) as an example, to mitigate the impact of industry differences and company sizes on the magnitude of the data, this study normalized the company market values, calculating the greenhouse gas emission intensity per unit of million Hong Kong dollar market value, thereby rendering the emission data of different companies comparable.

\begin{figure*}[!t]
\centering
\includegraphics[width=.95\linewidth]{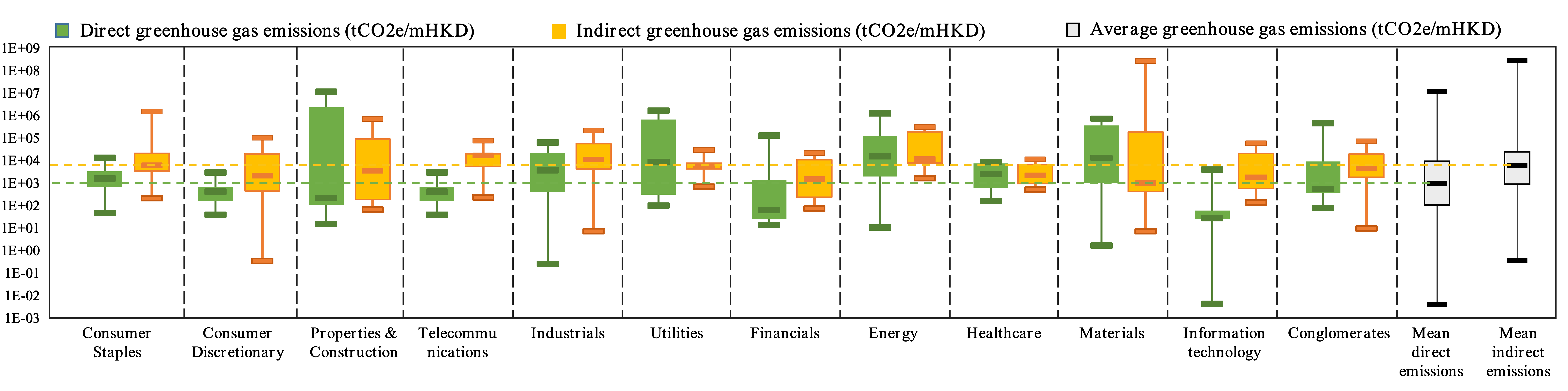}
\caption{\textrm{Comparative Analysis of Greenhouse Gas Emission Intensity for Direct and Indirect Emissions by Industry.}}
\label{fig:5}
\end{figure*}

According to the data displayed in Figure \ref{fig:5}, we observed a general trend: in most industries, the average Scope 2 emissions are higher than Scope 1 emissions. However, there are some exceptions, for example, in the Utilities, Energy, Healthcare and Materials, the level of Scope 1 emissions exceeded that of Scope 2. The Utilities sector, in particular, is prominent in Scope 1 emission data due to its coverage of energy-intensive large enterprises such as natural gas, coal gas, and electricity. In contrast, the Financials sector, as a typical labor-intensive industry, is more prominent in Scope 2 emission data. This trend, corroborated by the industry's data on electricity consumption, highlights the significant reliance of the Financials industry on electricity.

\subsubsection{Key Actions}
Based on the analysis of key action data extracted from ESG reports, we further summarized and generalized this data to reveal core information within the text. Supplementary Figure \ref{supplefig:2} displays the top five high-frequency words that we extracted from cross-industry environmental issue discussions. For instance, within the theme of greenhouse gas emission management, "Greenhouse gas reduction" is the most frequently mentioned phrase across industries, with a word frequency of 24.3\%. For water resource management, "Adopt water-saving practices" and "Water-saving training for staff" are equally frequent, each surpassing 10\%. Discussions related to the use of fossil fuels indicate a gradual shift from fossil fuels toward green, renewable energy, as evidenced by terms like "Renewable energy" and "Green transition". Dialogues on hazardous and non-hazardous waste management highlight "Pollution source control" and "Compliant emissions". Additionally, the topics of electricity and energy usage emphasize "Green office practices" and "Smart energy control". Overall, through a comprehensive analysis of high-frequency vocabulary, we find that expressions such as "ESG management" and "Green transition" frequently appear in ESG reports from different industries. These results reflect common actions across industries to reduce environmental impact.

Supplementary Table \ref{suppletab:5} lists key action words that are closely associated with specific industries, reflecting industry-specific action tendencies and dependencies. For instance, in Properties \& Construction sector, including companies such as China National Building Material Company Ltd. and China Resources Cement Holdings Ltd., there is an emphasis on controlling dust and volatile organic compound emissions, implementing water-saving systems, and reusing building materials. In Telecommunications sector, companies including China Tower Corporation Ltd. and China Mobile Ltd. focus on the research and development of environmentally friendly products, advancing the recycling and reuse of batteries, and enhancing environmental monitoring and pollution control related to communication towers. As for Healthcare sector, companies such as China Biologic Products Holdings, Inc. and Alibaba Health Information Technology Ltd. are concentrating their ESG efforts on promoting the development of digital medical services. Financials sector plays a leading role in the field of green finance, utilizing financial instruments to support the sustainable development of industries. 

\section{Discussion}
Although this study has yielded some initial findings, we have identified several factors that suggest avenues for further research. First, we have noted substantial differences in the capabilities of various LLMs when it comes to ESG analysis tasks. These differences underscore the reliance of LLMs on their intrinsic computational abilities and the depth of their ESG domain knowledge when performing specialized analyses. Second, the structural parsing and retrieval accuracy of ESG reporting documents are of paramount importance when LLMs are employed for ESG disclosure analysis. Enhanced document structuring and more effective information retrieval can notably improve the performance of the RAG framework in terms of data extraction and analysis. Lastly, given that key information in ESG reports is sometimes presented in pictorial forms, our current ESGReveal approach falls short in extracting such data. We aim to rectify this in our future work by refining and optimizing our approach to accommodate these kinds of information.

Looking to the future, research in this domain can evolve in several promising directions. One key direction is the meticulous development of more granular, industry-specific ESG datasets. Such datasets can drive highly accurate data analyses, significantly boosting the transparency and credibility of corporate ESG disclosures. Moreover, the automated extraction techniques pioneered in this study are anticipated to broaden their reach, with potential applications extending to the analysis of pivotal documents such as those from the Intergovernmental Panel on Climate Change. This expansion would leverage digital technology to make a meaningful impact on critical global conversations around climate change mitigation and sustainable development.

In summary, the continued refinement of technological tools and frameworks, along with the expansion of the database's scope of application, promises to not only furnish businesses with invaluable insights but also to play a part in steering entire industries towards a path of enhanced transparency and sustainability.

\section{Conclusion}
In the domain of ESG reporting analysis, this study introduced a method for data extraction, ESGReveal, suited to a multitude of ESG compilation standards. The method is anchored in the RAG paradigm and leverages LLM to accomplish structured extraction of disclosed data within ESG reports. Furthermore, this study utilized the 2022 ESG reports from a representative subset of companies listed on the HKEx to build a structured database for each industry and conducted a quantitative analysis. The pivotal findings of this study are as follows:

(1) The significance of ESG metadata module in ESG Analysis.

The ESG metadata module, rooted in LLM technology, enables the conversion of various ESG reporting standards into structured data extraction commands. By defining ESG indicator attributes and providing an extension mechanism, it bolsters adaptability in a multi-standard landscape. The module significantly enhances LLM performance in ESG analytical tasks, evidenced by gains of +9.9\% in GPT-3.5 and +2.5\% in GPT-4.

(2) The Applicability of ESGReveal in ESG Analysis Tasks.

ESGReveal, built upon LLM and the RAG paradigm, enables efficient extraction of key indicator data and essential actions from ESG reports. On GPT-4, the accuracy rates for data extraction and disclosure analysis tasks reached 76.9\% and 83.7\%, respectively, representing an increase of over 20\% compared to baseline tests. These results amply demonstrate the potent potential and practical application value of ESGReveal in ESG analysis tasks.

(3) Cross-Industry ESG Data Disclosure and Key Action Analysis at HKEx.

Upon analyzing ESG data disclosures of representative companies under the HKEx framework, the findings reveal that the average disclosure rate for environmental data stands at 69.5\%, while social data disclosure averages at 57.2\%. The overall disclosure rate does not exceed 80\%, indicating a need for enhanced ESG practices across all sectors. Furthermore, an analysis of key action terms extracted has identified common ESG actions implemented across industries, as well as unique actions that are characteristic of specific sectors.



\bibliographystyle{cas-model2-names}

\bibliography{ref.bib}



\section*{Supplementary Material}

\setcounter{table}{0}
\begin{table*}[ht]
\centering
\caption{Indicators of the ESG Metadata Module Based on HKEx Standards.}
\label{suppletab:5}
\includegraphics[height=0.95\textheight]{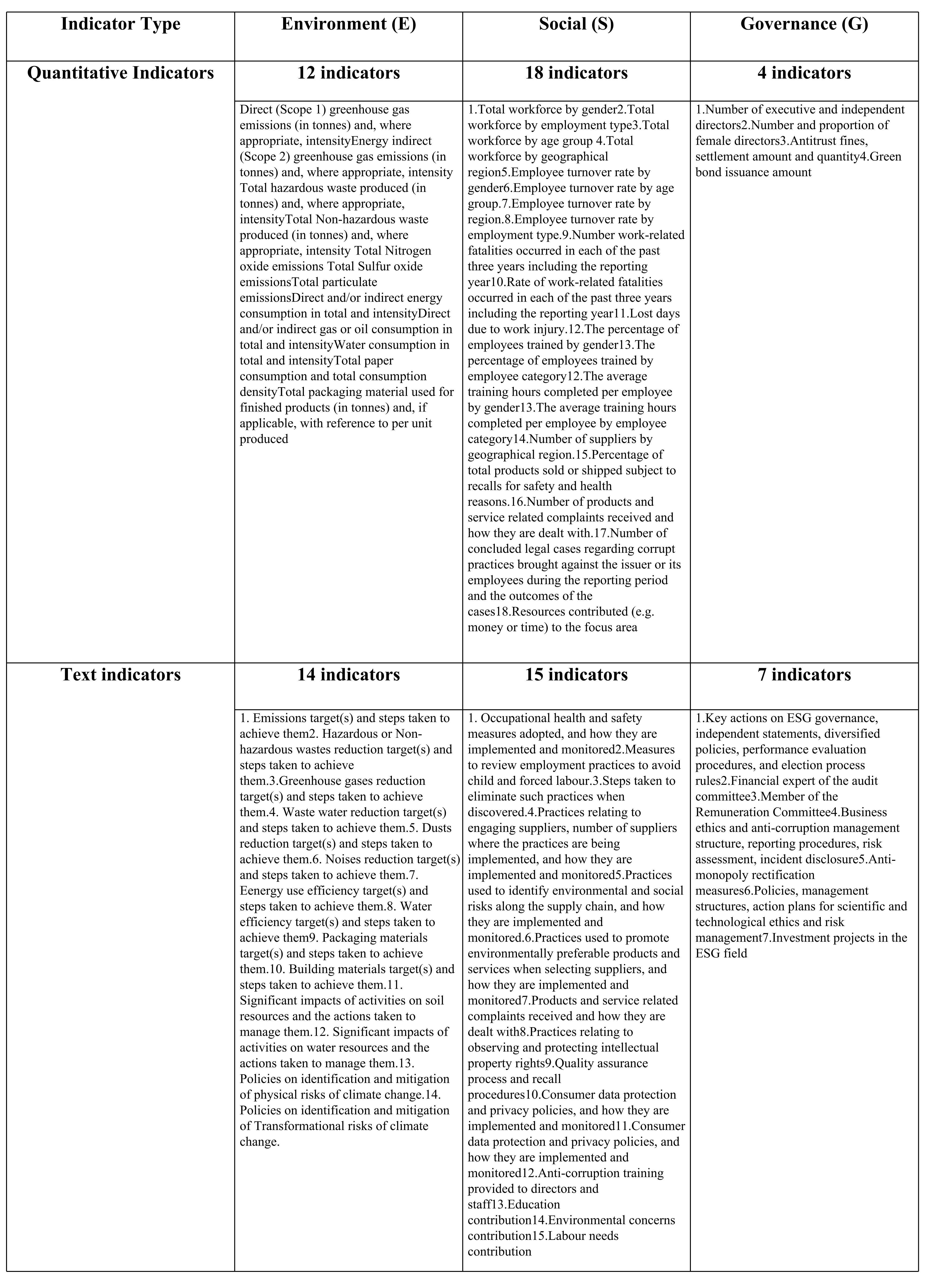}
\end{table*}

\clearpage 

\setcounter{table}{1}
\begin{table*}[ht]
\centering
\caption{Examples of the ESG Metadata Module Based on HKEx Standards.}
\label{suppletab:5}
\includegraphics[height=0.95\textheight]{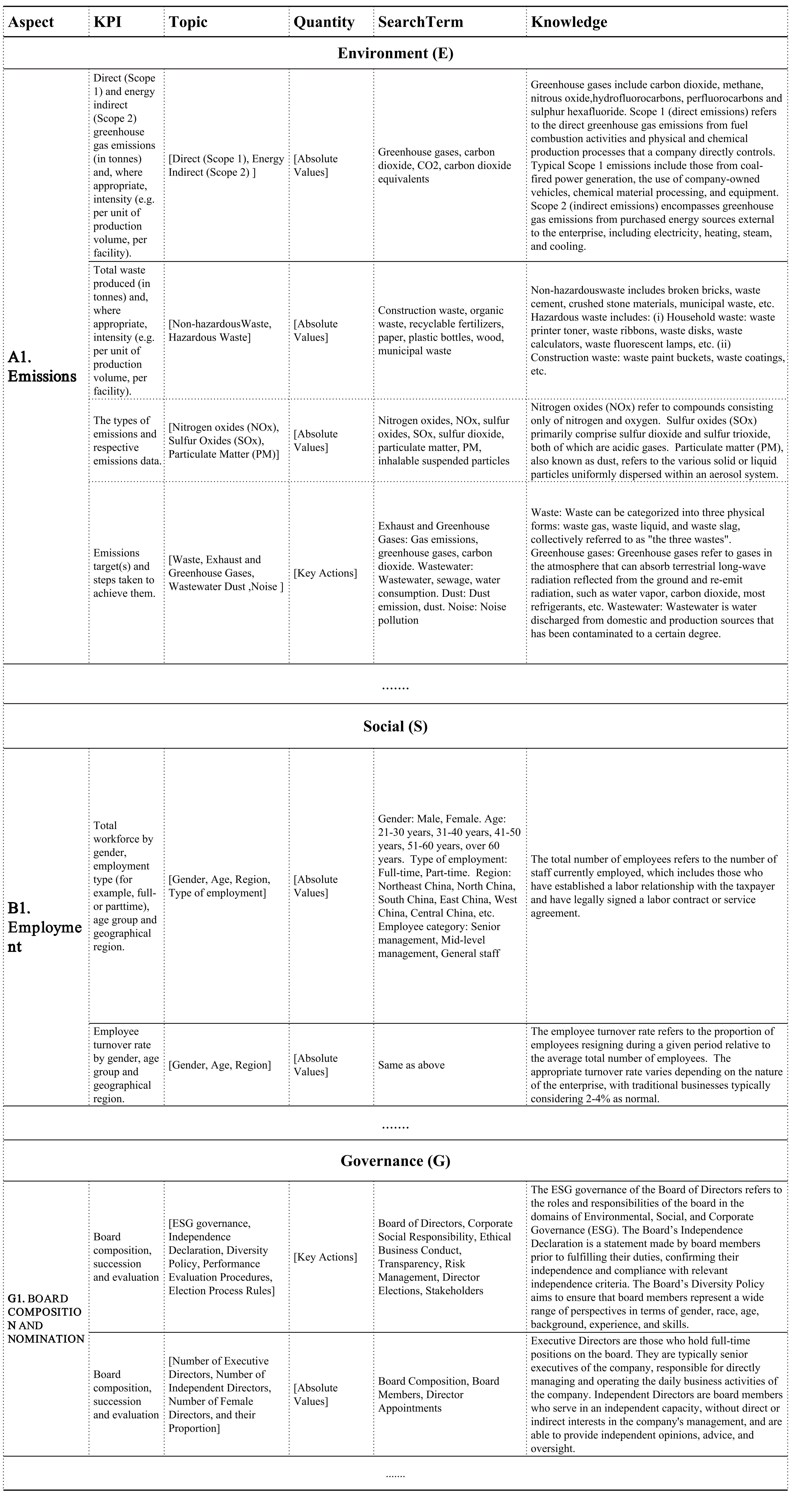}
\end{table*}

\clearpage 

\setcounter{table}{2}
\begin{table}[ht]
\centering
\caption{Examples of Prompts and Answers.}
\label{suppletab:3}
\includegraphics[height=0.95\textheight]{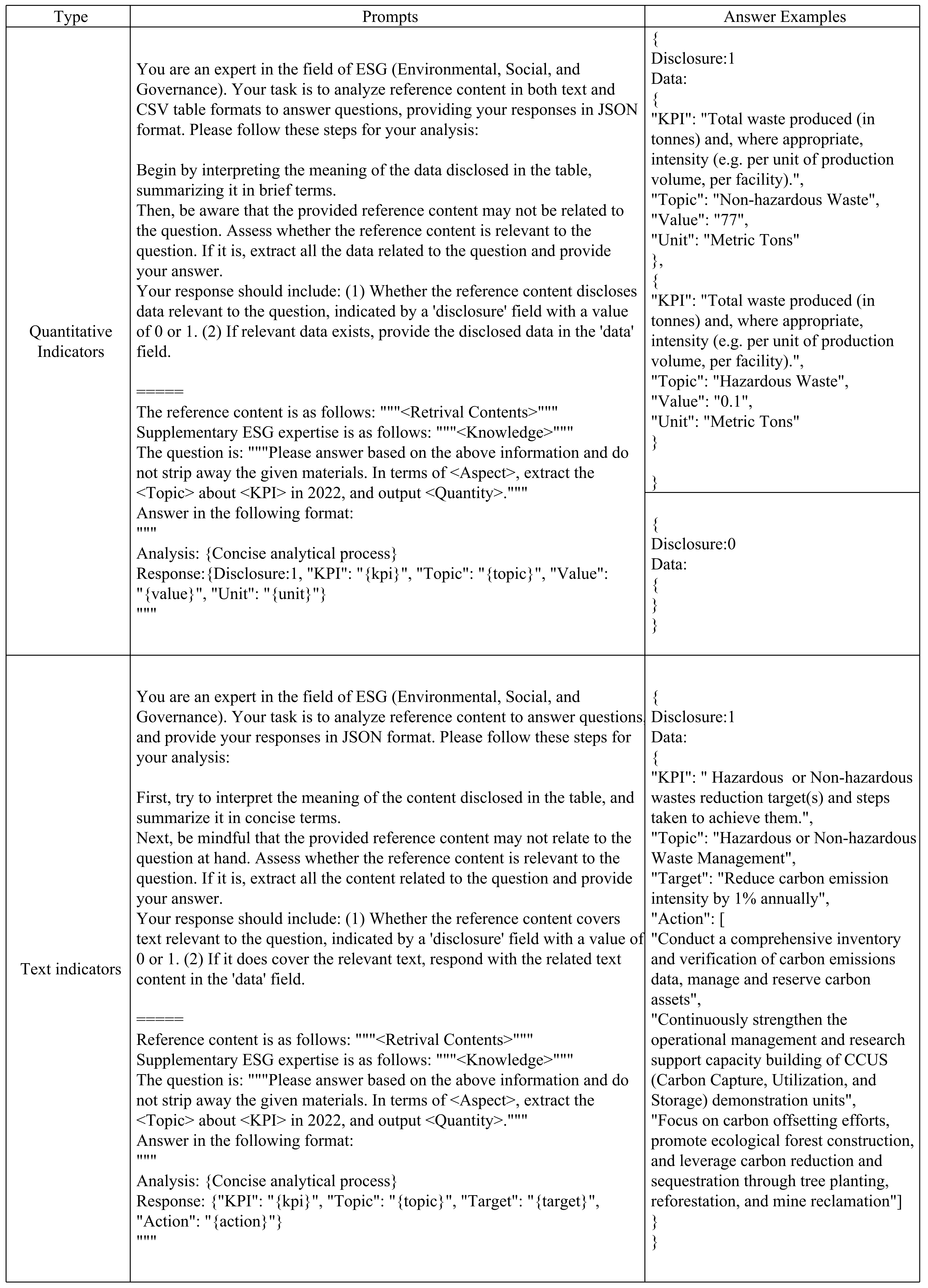}
\end{table}

\clearpage 

\setcounter{table}{3}
\begin{table*}[ht]
\centering
\caption{Representative Companies in Each Sector According to the Hang Seng Industry Classification System.}
\label{suppletab:4}
\includegraphics[height=0.95\textheight]{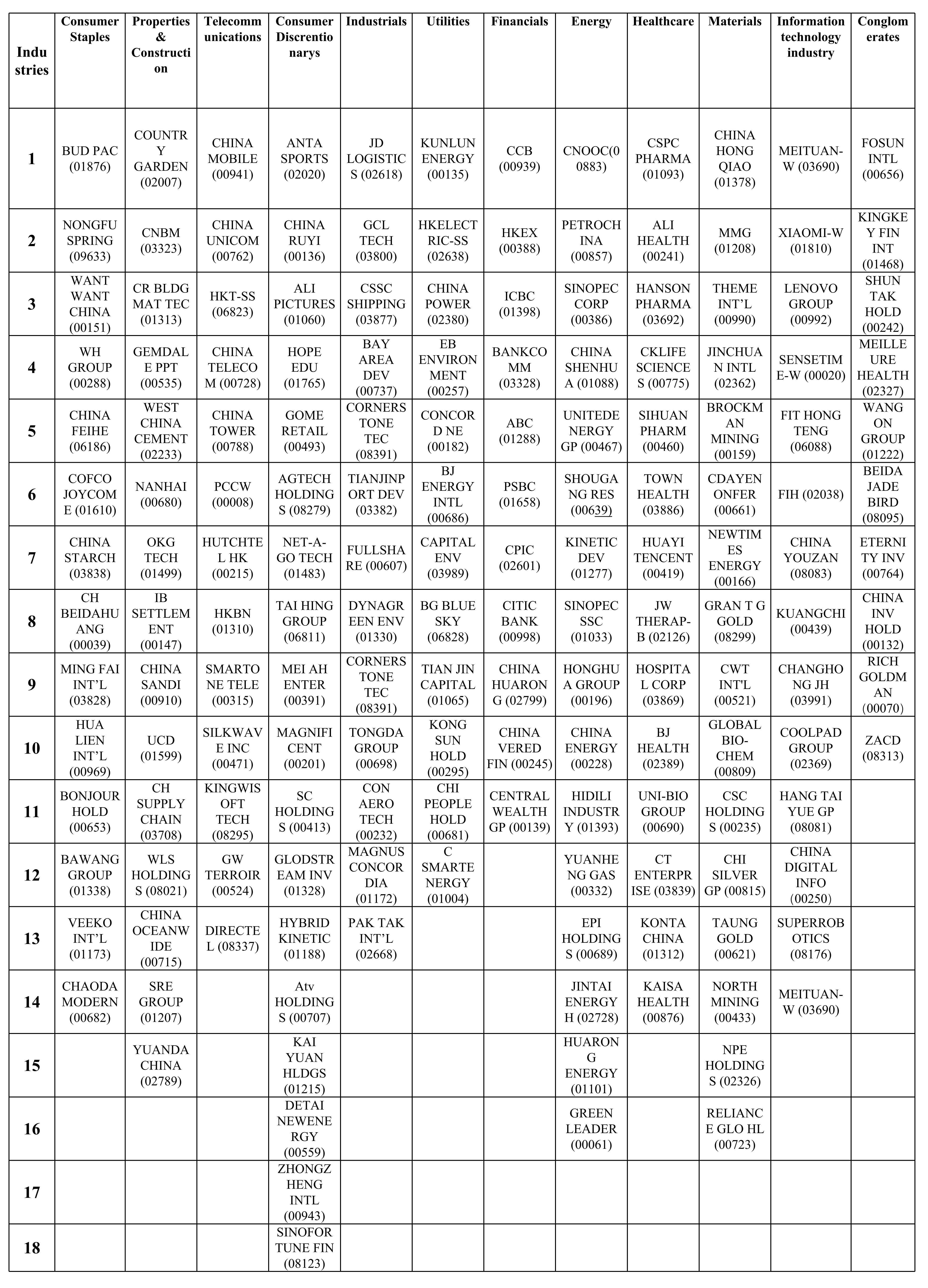}
\end{table*}

\clearpage 

\setcounter{figure}{0}
\begin{figure*}[!t]
\centering
\includegraphics[width=.95\linewidth]{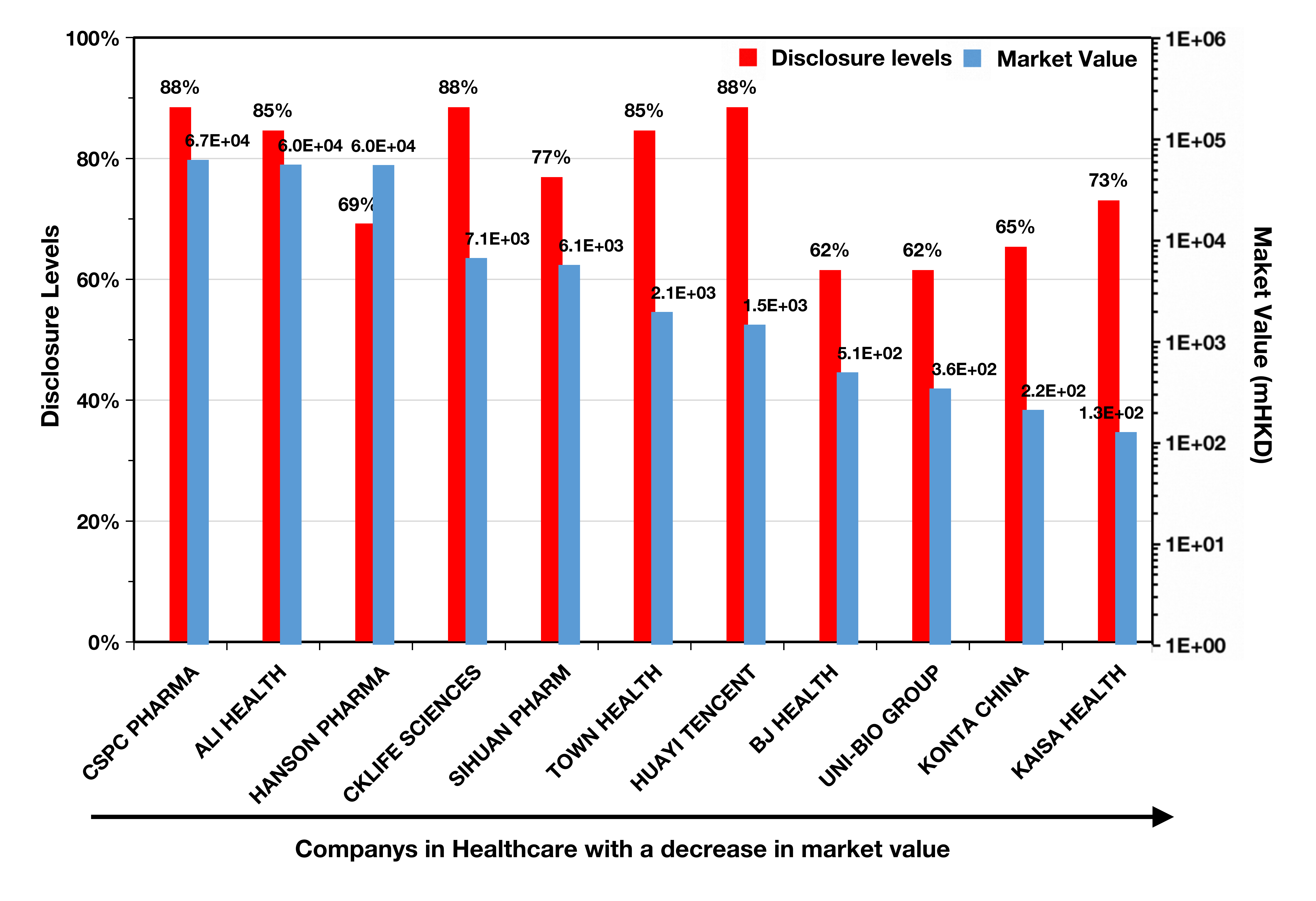}
\caption{\textrm{Disclosure Levels and Market Value Across Different Companies in Healthcare.}}
\label{supplefig:1}
\end{figure*}

\clearpage 

\setcounter{figure}{1}
\begin{figure*}[!t]
\centering
\includegraphics[width=.95\linewidth]{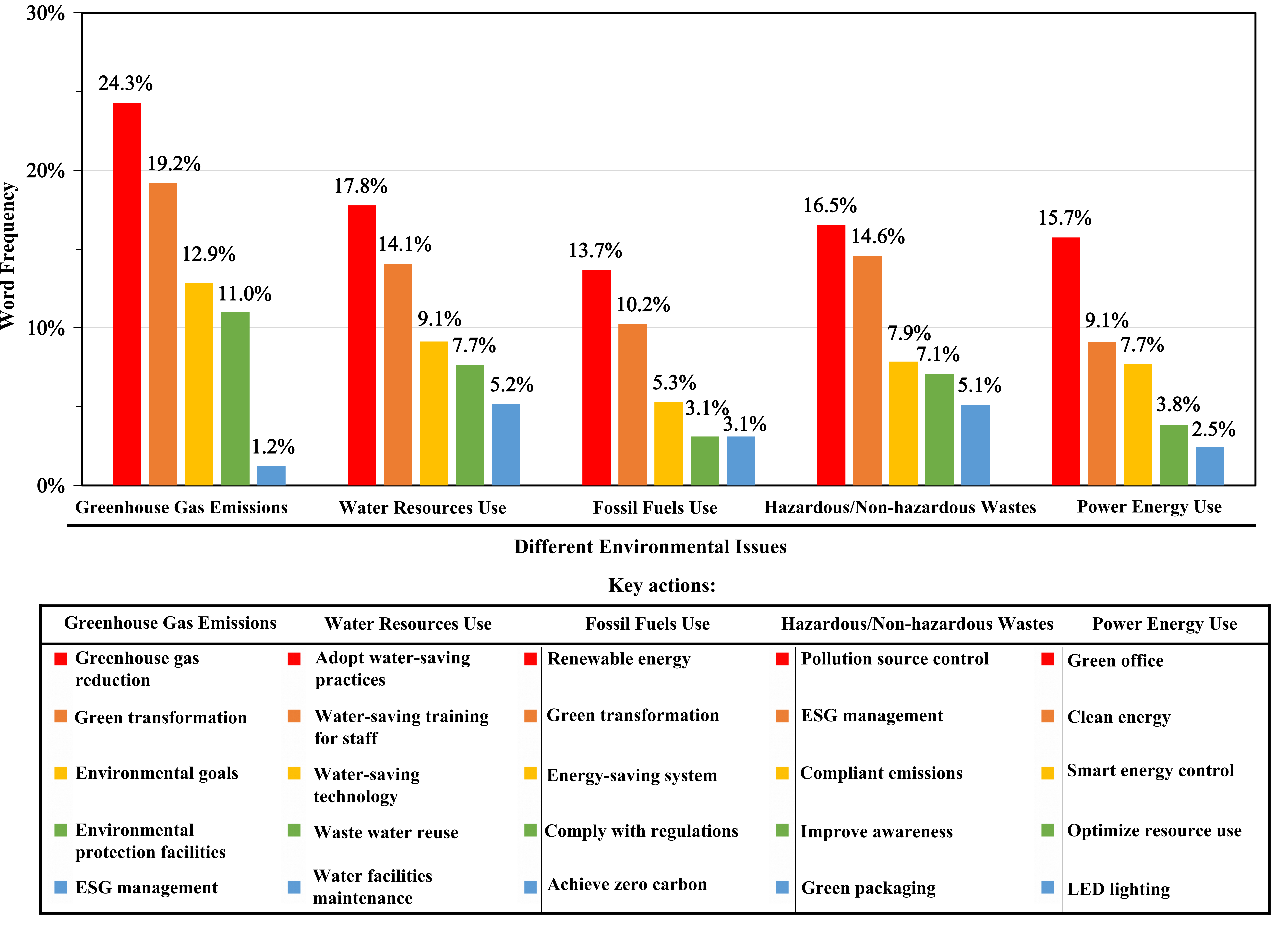}
\caption{\textrm{Key Actions and Word Frequencies Under Different Environmental Issues.}}
\label{supplefig:2}
\end{figure*}

\clearpage 

\setcounter{table}{4}
\begin{table*}[ht]
\centering
\caption{Partial Examples of Key Actions by Industries for Different Environmental Issues.}
\includegraphics[height=0.95\textheight]{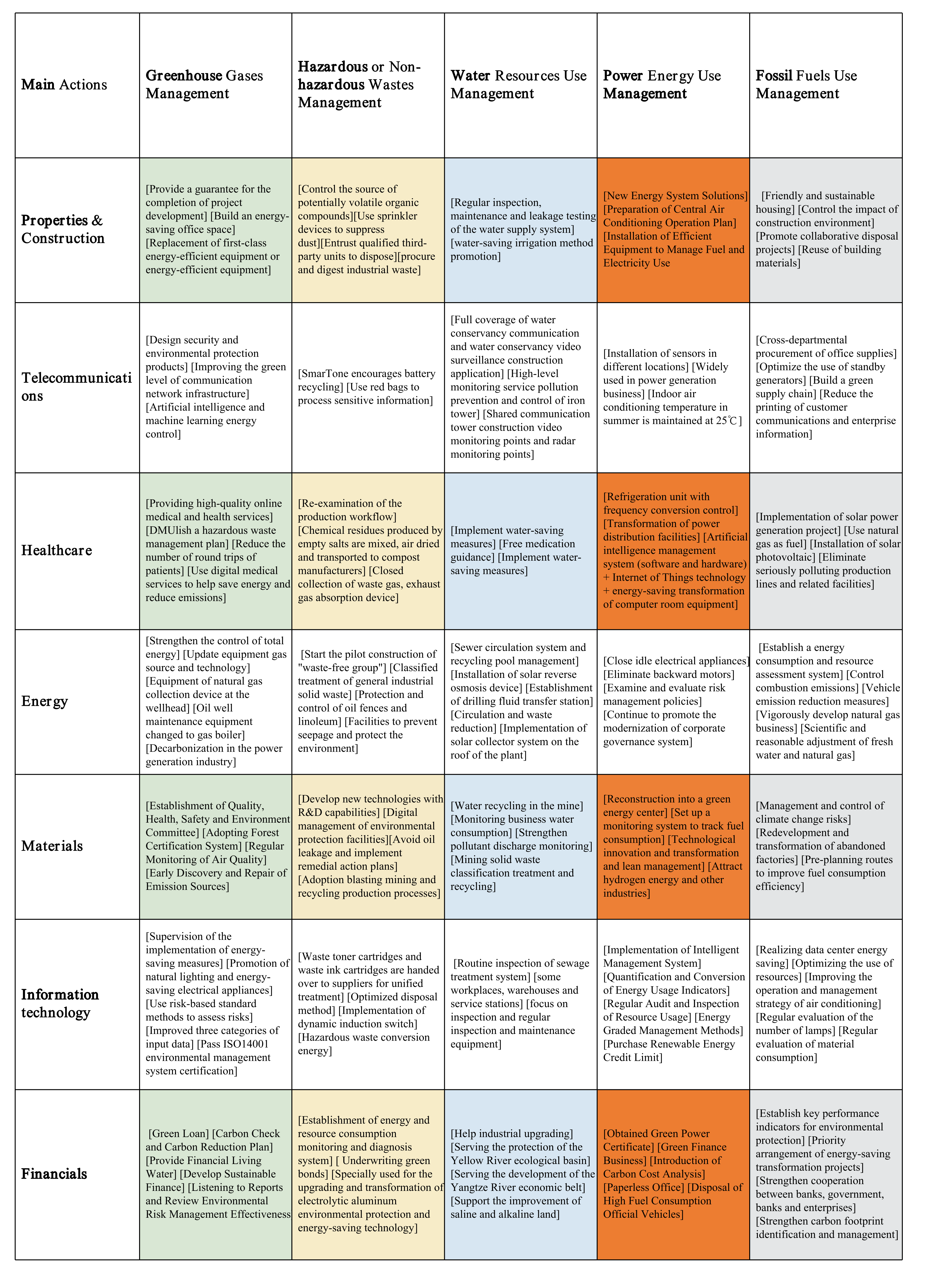}
\end{table*}

\end{document}